\documentclass[10pt,twocolumn,letterpaper]{article}

\usepackage[pagenumbers]{cvpr} % To force page numbers, e.g. for an arXiv version
\usepackage{graphicx}
\usepackage{amsmath}
\usepackage{amssymb}
\usepackage{mathtools}
\usepackage{booktabs}
\usepackage{adjustbox}
\usepackage{multicol, multirow}
\usepackage{float}
\makeatletter
\@namedef{ver@everyshi.sty}{}
\makeatother
\usepackage{tikz}
\usepackage{epsfig, svg}
\usepackage{caption, subcaption}
\usepackage{amsthm}
\theoremstyle{plain}
\newtheorem{theorem}{Theorem}[section]

\newtheorem{corollary}[theorem]{Corollary}
\newtheorem{observation}[theorem]{Observation}
\newtheorem{proposition}[theorem]{Proposition}

\theoremstyle{definition}
\newtheorem{definition}[theorem]{Definition}

\theoremstyle{remark}

\newcommand{\reals}{\mathbb{R}}

\newcommand{\integers}{\mathbb{Z}}
\newcommand{\norm}[1]{\left\lVert #1 \right\rVert}
\newcommand{\innprod}[2]{\langle #1,#2 \rangle}
\newcommand{\intunitsq}{\int_{-\frac{1}{2}}^{\frac{1}{2}}}

\newcommand{\openzinter}[3]{\left(#1, #2\right)^{#3}_\integers}
\newcommand{\relu}{\text{\rm{ReLu} }}
\newcommand{\trm}[1]{\text{\rm{#1}}}
\usepackage{acronym}

\acrodef{AT}[AT]{Adversarial Training}
\acrodef{FGSM}[FGSM]{Fast Gradient Sign Method}
\acrodef{PGD}{Projected Gradient Descent}
\acrodef{FGSM-AT}[FGSM-AT]{FGSM Adversarial Training}
\acrodef{PGD-AT}[PGD-AT]{PGD Adversarial Training}
\acrodef{CW}[CW]{Carlini-Wagner}

\acrodef{CNN}[CNN]{Convolutional Neural Network}
\acrodef{CNNs}[CNNs]{Convolutional Neural Networks}
\acrodef{GAN}[GAN]{Generative Adversarial Network}

\acrodef{BP}[BP]{BlurPool}
\acrodef{MBP}[MBP]{MaxBlurPool}
\acrodef{BMP}[BMP]{BlurMaxPool}
\acrodef{BMBP}[BMBP]{BlurMaxBlurPool}
\acrodef{BAP}[BAP]{BlurApplyPool}
\acrodef{ABP}[ABP]{ApplyBlurPool}
\acrodef{BABP}[BABP]{BlurApplyBlurPool}
\acrodef{QA}[QA]{Quantile Adjustment}
\acrodef{PO}[PO]{Permanent Over-sampling}
\acrodef{FA}[FA]{Fully Anti-aliased}
\acrodef{PA}[PA]{Partially Anti-aliased}

\acrodef{FS}[FS]{Fourier Series}
\acrodef{DFT}[DFT]{Discrete Fourier Transform}
\acrodef{SIFT}[SIFT]{scale-invariant feature transform}

\acrodef{UAC}[UAC]{Uniform Absolute-Convergence}
\usepackage[pagebackref,breaklinks,colorlinks]{hyperref}

\usepackage[capitalize]{cleveref}
\crefname{section}{Sec.}{Secs.}
\Crefname{section}{Section}{Sections}
\Crefname{table}{Table}{Tables}
\crefname{table}{Tab.}{Tabs.}

\def\cvprPaperID{9953} % *** Enter the CVPR Paper ID here
\def\confName{CVPR}
\def\confYear{2023}

\begin{document}

%%%%%%%%% TITLE - PLEASE UPDATE
\title{Aliasing is a Driver of Adversarial Attacks}

\author{Adrián Rodríguez-Muñoz\\
MIT CSAIL\\
{\tt\small adrianrm@mit.edu}
% For a paper whose authors are all at the same institution,
% omit the following lines up until the closing ``}''.
% Additional authors and addresses can be added with ``\and'',
% just like the second author.
% To save space, use either the email address or home page, not both
\and
Antonio Torralba\\
MIT CSAIL\\
{\tt\small torralba@mit.edu}
}
\maketitle

%%%%%%%%% ABSTRACT
\begin{abstract}
Aliasing is a highly important concept in signal processing, as careful consideration of resolution changes is essential in ensuring transmission and processing quality of audio, image, and video. Despite this, up until recently aliasing has received very little consideration in Deep Learning, with all common architectures carelessly sub-sampling without considering aliasing effects. In this work, we investigate the hypothesis that the existence of adversarial perturbations is due in part to aliasing in neural networks. Our ultimate goal is to increase robustness against adversarial attacks using explainable, non-trained, structural changes only, derived from aliasing first principles. Our contributions are the following. First, we establish a sufficient condition for no aliasing for general image transformations. Next, we study sources of aliasing in common neural network layers, and derive simple modifications from first principles to eliminate or reduce it. Lastly, our experimental results show a solid link between anti-aliasing and adversarial attacks. Simply reducing aliasing already results in more robust classifiers, and combining anti-aliasing with robust training out-performs solo robust training on $L_2$ attacks with none or minimal losses in performance on $L_{\infty}$ attacks. %Reducing aliasing results in more robust classifiers, that while not up to the level of benchmark methods, do not require robust training and for whom the reason behind their robustness is understood. Furthermore, combining anti-aliasing with robust training out-performs solo robust training on $L_2$ attacks with only minimal losses on $L_{\infty}$ attacks.

\end{abstract}

%%%%%%%%% BODY TEXT
\section{Motivation}
\label{sec:motivation}

Deep Neural Networks (DNN) have become the state of the art in many different machine learning tasks. In particular, Deep Convolutional Networks have achieved near human-level accuracy in image classification challenges \cite{Krizhevsky2012, He2015}. However, many real-world applications require high standards of reliability, safety, and interpretability; in this sense, DNNs are not yet up to the task. One of the key reasons for this setback is the existence of adversarial examples, imperceptible perturbations to images that drastically change the predictions of neural networks with very high probability \cite{Szegedy2014, Goodfellow2015, He2015}.

There has been considerable work on developing defenses against adversarial examples \cite{Szegedy2014, Papernot2016, Buckman2018, Xu2018}, with \ac{AT} \cite{Goodfellow2015, Kurakin2017, Madry2019} standing out as the strongest current paradigm. However, there have also been significant advances towards developing more powerful methods of attack \cite{Szegedy2014, Goodfellow2015, MoosaviDezfooli2016, Carlini2017, Chen2018, Sharma2018}, such that the issue is as of yet far from solved.
Works characterizing adversarial examples in an analytic setting \cite{Ilyas2019, Tsipras2019} often do so as well-chosen but general points within a neighborhood of benign images. These type of approaches yield domain agnostic frameworks revolving around functional analysis concepts like Lipschitz continuity or estimator robustness. However, adversarial attacks are not just any kind of perturbation; they are often white-like noise that is imperceptible to humans. %While more human-noticeable perturbations of the same $L_\infty$ or $L_2$ norm conceivably exist, the fact that the most common avenue of attack for most adversarial algorithms is inexplicable by humans suggests that image processing by neural networks and human perception are somehow at odds. It is this incongruity that motivates the subject of our work.

\begin{figure*}[htbp]
\centering
\resizebox{.9\linewidth}{!}
{\includegraphics{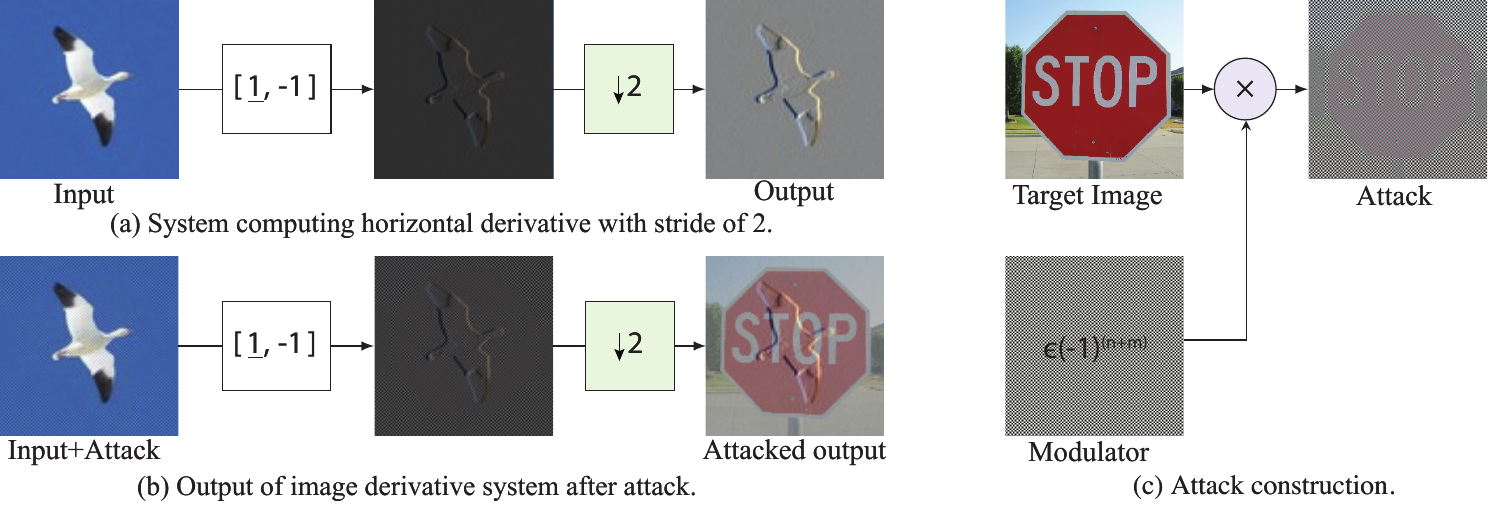}}
\caption{Analytic aliasing-based attack of a toy system that computes the horizontal image gradient, shown in (a). As this simple system has aliasing, an attacker, shown in (b), can inject a high frequency pattern to dramatically change the output with an imperceptible change to the input. The dynamic range of the images has been re-scaled to [0,1] for visualization purposes. In this example $\epsilon=\frac{16}{255}$.}
\label{fig:toy_example}
\end{figure*}

Our hypothesis is that adversarial attacks work in part by exploiting the phenomenon of aliasing. %, which relates to one such image processing incongruity: humans and machines process changes in resolution very differently. 
Formally, aliasing is a perceptual phenomenon whereby the appearance of a signal, visual or otherwise, can change drastically after sub-sampling. See \cref{fig:toy_example} for a simple but enlightening toy example. As we can see, the dirty image is indistinguishable from the original clean image by a human, and yet their outputs are completely different. The culprit for this bizarre effect is the convolution stride (green box in \cref{fig:toy_example}) that carelessly down-samples the input. An attacker with knowledge about it is able to construct a perturbation focused on manipulating the surviving samples (pixels at even rows and columns). The discarded samples (pixels at an odd row or column) serve as extra degrees of freedom that can be used to make the attack less noticeable and more powerful. The behavior of these analytically constructed attacks is remarkably similar to low-amplitude, gradient-driven attacks: they are imperceptible by humans and drastically change the feature maps of a network. It is thus plausible that attacks may be exploiting aliasing.

%-------------------------------------------------------------------------
\section{Related work}
\label{sec:related-work}

% Adversarial Attacks and Defences
%\subsection{Adversarial attacks and defenses}

{\bf Adversarial Attacks}, broadly speaking, can be split into two types: low-amplitude and large-amplitude attacks. {\em Low-amplitude attacks} enforce imperceptibility by adding a limited $L_p$ constraint to the adversarial noise. Historically, the most common choice has been $p=\infty$, but there is recent work that has called this decision into question \cite{Carlini2017, Carlini2017a, Chen2018} by showing that protection against $L_\infty$ attacks does not imply protection against $L_1$ or $L_2$. Well known attacks of this type are FGSM ($L_\infty$) \cite{Goodfellow2015}, DeepFool ($L_2$) \cite{MoosaviDezfooli2016}, PGD ($L_\infty, L_2$) \cite{Madry2019}, CW ($L_2$) \cite{Carlini2017}, EAD ($L_1 + L_2$) \cite{Chen2018}, and DI-FGSM ($L_\infty$) \cite{Xie2019}. {\em Large-amplitude attacks} discard the $L_p$ constraint and instead leverage knowledge of the human visual system to produce imperceptible adversarial noises. For example, STA \cite{Xiao2018} slightly alters the shape of MNIST \cite{MNIST} digits in a way that fools classifiers, yet evades human detection. Another interesting work in this area involves attacking the scaling algorithms that precede most Neural Classifiers, as done in \cite{Xiao2019}. The attack is independent of the classifier architecture and requires parallel work on robust scaling algorithms to combat as in Quiring \etal \cite{Quiring2020}. The small transformations considered in \cite{Zhang2019} are also imperceptible yet radically change predictions.

{\bf Adversarial Defenses} can also be split into two types: robustness-based and detection-based defenses. {\em Robustness-based defenses} aim to produce models that correctly classify adversarial examples. The current strongest paradigm for robustness-based defenses is \ac{AT} \cite{Goodfellow2015, Kurakin2017, Madry2019}. While originally it seemed that \ac{AT} had an essential trade-off between clean accuracy and robustness, recent work has produced approaches where this no longer seems to be the case \cite{Xie2020, Poursaeed2021}. {\em Detection based defenses} \cite{Xu2018} produce models that are able to differentiate between clean and attacked examples, and refuse to classify in the latter case. However, \cite{Carlini2017a} and \cite{Sharma2018} have largely defeated these approaches.

For a detailed extensive review of work done in adversarial attacks and defenses see \cite{Xu2019}.

{\bf Aliasing in Deep Neural Networks} has been a largely ignored topic until recent years. \cite{Zhang2019} presented the issue of aliasing in max-pool and strided convolution layers in terms of (lack of) translation invariance. \cite{Chaman2020} solved this problem entirely in devising a clever and simple trick by which they choose the phase of the sub-sampling via an energy criterion, rather than always using 0-phase, such that sub-samplings are now perfectly unit-shift invariant. \cite{Zou2020} expanded on classic anti-aliasing blurring by introducing location and channel-dependent blurring. \cite{Hossain2021} took the matter further and treated aliasing in non-linearities, as well as enforced progressively stronger anti-aliasing \wrt{} depth. \cite{Karras2021} brought the concept into GANs \cite{Goodfellow2014}, and proved how aliasing was responsible for the detail coordinate-sticking effect. \cite{Grabinski2021} proved the existence of aliasing in the down-sampling step of \ac{CNNs}, and, most remarkably, made the acute observation that aliasing coincided with adversarial vulnerability. Similarly, \cite{Tsuzuku2018,Yin2019,Chan2022} showed the relevance of Fourier analysis in robustness and adversarial attacks. \cite{Vasconcelos2021} made an extensive study into the placement of blurring kernels and the benefits of explicit untrained anti-aliasing on generalization. \cite{Ribeiro2021} showed that while CNNs are capable of distinguishing oscillations and, in principle, implementing anti-aliasing blurring, this does not prevent aliasing from taking place.

To the best of our knowledge, we are the first to propose concrete anti-aliasing approaches as an untrained structural defense to white-box attacks. This is a completely different scope than in \cite{Chaman2020,Vasconcelos2021}, which focus on the clean generalization benefits of anti-aliasing. \cite{Grabinski2021} also arrived at the hypothesis that aliasing is an underlying cause of adversarial vulnerability, but did not propose defenses.

We differ from previous work in a three-fold way. First, we expand on blurring-based anti-aliasing approaches like in \cite{Zhang2019,Vasconcelos2021} by theoretically deriving the appropriate blurring strength. Second, we propose the completely novel Quantile ReLu, an anti-aliasing modification that is independent of blurring-based approaches. As we will see in the results section, combining blurring with the quantile modification yields additive robustness gains and is essential to obtain robustness. Lastly, our experiments empirically show that the anti-aliasing modifications we propose already serve as natural defenses that significantly increase robustness to any-amplitude single-step attacks and low-amplitude multi-step attacks. Moreover, combining anti-aliasing with robust training out-performs solo robust training on $L_2$ attacks of all amplitudes with no or minimal losses in performance on $L_{\infty}$ attacks.

%-------------------------------------------------------------------------
\section{Neural Networks without aliasing}
\label{sec:methodology}

Neural networks are full of aliasing. In this section, we briefly explain the concept of aliasing and establish a general approach to anti-aliasing arbitrary image transformations, which we then apply to the specific transformations found in \ac{CNNs}. \cref{figure:layercards} shows a summary graphic of the adaptations used.

To combat aliasing, we expand on the existing blurring-based approaches such as in \cite{Zhang2019,Vasconcelos2021,Karras2021} by using theory to derive the exact blurring strength necessary, which coincides with the experimentally derived strength as was done in \cite{Karras2021}. Furthermore, we also introduce the Quantile ReLu anti-aliasing modification, which is a new way of anti-aliasing independent of and synergistic with blurring-based approaches. %As we will see in the results section, combining filtering methods and the quantile modification yields additive robustness gains and is essential to obtain robustness.

\subsection{What is aliasing?}

The concept of aliasing is intrinsically related to discrete sampling. In layman's terms, the more "complex" a continuous-domain signal, the finer the sampling needed to properly represent it. Using an insufficiently fine sampling results in visual artifacts that perceptually destroy the original signal; we call this phenomenon "aliasing".

\begin{figure}
	\centering
    %\resizebox{\linewidth}{!}{\input{figures/aliasing_example/main.tikz}}
    \includegraphics[width=1\linewidth]{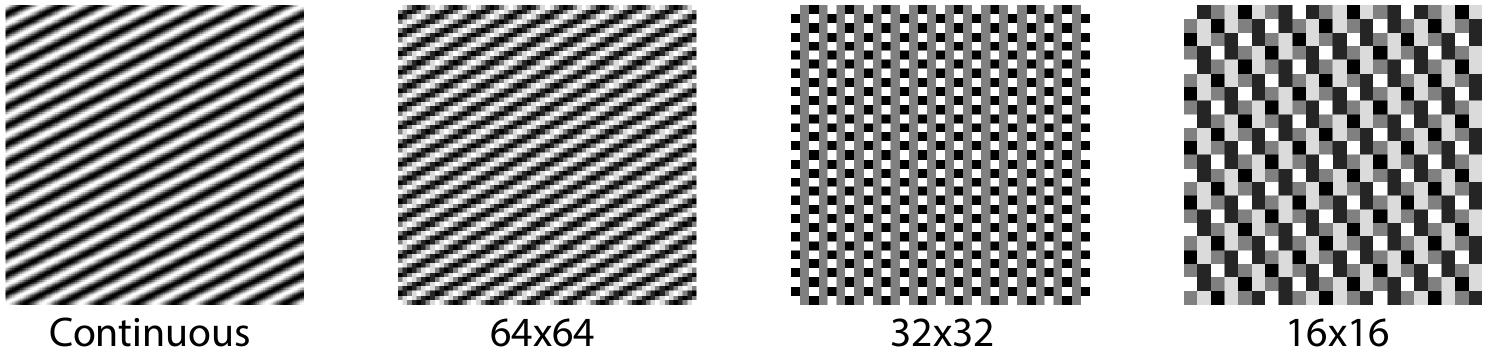}
    \caption{An insufficiently dense sampling of
    %\begin{equation}
    %    z(x,y)=\cos\left(2\pi4\left(x+y+\frac{8}{16}\right)\right)
    %\end{equation}
    %$\cos\left(8\pi\left(x+y+0.5\right)\right)$
    $\cos\left(16\pi\left(x+2y\right)\right)$
    with resolution $16\times16$ leads to disastrous aliasing, as the main feature of the signal, the bottom-left to top-right diagonals, is inverted.}
    \label{fig:aliasing_diagonals_example}
\end{figure}

Consider the example shown in \cref{fig:aliasing_diagonals_example}: the main "feature" of the signal, the right-to-left diagonals, is inverted by aliasing when sampling at an insufficient rate. Our hypothesis is that such visual artifacts in the processing of a \ac{CNN} could be leveraged by attacks to confound networks, as motivated by \cref{fig:toy_example}, by providing a mechanism by which a seemingly innocuous signal may drastically change during processing.

The Shannon-Nyquist sampling theorem formalizes the concept of image complexity and provides the necessary sampling rate:
\begin{theorem}[Shannon-Nyquist sampling theorem]
\label{theorem:shannon_nyquist_sampling_theorem}
A continuous-domain 1-periodic signal $z$ is uniquely represented by a sampling with rate $s$ if and only if its \ac{FS} contains no non-zero terms for frequencies greater than or equal to $s/2$ \cite{ClaudeShannon1949}. In the positive, we call the continuous-domain signal "representable" and the sampling "valid". The greatest non-zero term frequency is called the band-limit of $z$.
\end{theorem}

We can apply the sampling theorem to derive a general approach. Given a transformation $T$ acting on discrete-domain inputs of resolution $s\times s$, we need only: (1) define $T$ for continous-domain inputs in a consisent manner, and (2) adapt $T$ such that it transforms representable continuous-domain inputs into representable continuous-domain outputs. In practical terms this will mean limiting the high-frequency information that $T$ can create, increasing the sampling rate $s$, or a combination of both.

There are three transformations in \ac{CNNs} that can generate aliasing: ReLu non-linearities, sub-sampling layers, and max-pool layers. Batch-Normalizations and (unstrided) convolutional layers do not cause aliasing due to being linear and affine transformations respectively that do not change the sampling rate.

\subsection{Reducing aliasing in the ReLu}
\label{sec:reducing-aliasing-in-the-relu}

It can be proven that point-wise polynomial transformations at most increase the band-limit by a factor equal to their degree. This means that up-sampling the feature map by a factor $U$ and blurring to a normalized frequency $\frac{1}{2L}$ lets us compute a polynomial non-linearity without aliasing iff $UL\geq\text{degree}$. For non-linearities that are not polynomials, but that are well-approximated by a small-degree polynomial on the distribution of the signal, we may suppress aliasing to a great degree by anti-aliasing as if we were computing the polynomial approximation. In the supp. mat., we rigorously prove that the aliasing error is bounded by twice the approximation error. Of course, what constitutes a good approximation and an acceptable aliasing error depends on the application.

In particular, the ReLu function can be approximated with less than 1\% MSE on standard normal distributed inputs by the fourth degree polynomial:
\begin{equation}
   \sum_{k=0}^4 \innprod{\trm{He}_k}{\relu}_{\mathcal{N}(0,1)}\trm{He}_k(t) = \frac{t}{2}-\frac{t^{4}-18t^{2}-9}{24\sqrt{2\pi}}
\end{equation}
which we compute using the Hermite polynomials $\{\trm{He}_k\}$ and the orthogonal projection theorem. This suggests that we may compute the ReLu with a small amount of aliasing via a combination of up-sampling the feature map by a factor $U$ and blurring to a normalized frequency $\frac{1}{2L}$ for any $U,L$ with $UL \geq 4$. This theory derived conjecture coincides with the practical results obtained in \cite{Karras2021} in a \ac{GAN} setting, and we use their efficient CUDA implementation in our experiments. %Given the appropriate filtering strength, this anti-aliasing modification is equivalent to the post-filtering of \cite{Vasconcelos2021}.

Unfortunately, it appears that blurring alone is insufficient to prevent aliasing in an adversarial setting, where extreme cases are the norm rather than an oddity. Given an input with a very strong high-frequency bias, a small absolute aliasing error may be a large relative aliasing error with respect to the low-frequency un-aliased signal.

To correct this situation, we take the following complementary approach which we apply in tandem with the blurring-type modification of \cite{Karras2021}. A ReLu non-linearity can be thought of as a non-uniform sampling that keeps positive samples and discards negative samples. Thus, we can reduce the relative aliasing error produced by enforcing a floor on the \% of samples that "survive", as this ensures an upper bound on how much the ReLu operation can change our signal. We call this \ac{QA}.

\begin{definition}[Quantile Adjusted ReLu]
\label{defintion:qrelu}
	The Quantile Adjusted ReLu of quantile $q$ is
	\begin{equation}
	\label{equation:qrelu}
	\text{QReLu}_q(Z) \coloneqq \max(Z + \max(-Z_q, 0), 0)
	\end{equation}
	where $Z_q$ is the $q$-th quantile of $Z$. In the context of a neural net the quantile is computed channel-wise. %See figure \ref{figure:QReLu} for a visual representation.
\end{definition}
In our experiments, we chose to take an aggressive approach and set $q=0.4$, which ensures that 60\% of the signal remains unchanged (up to a fixed per-channel additive shift). 

The quantile ReLu is completely different from previous anti-aliasing approaches like in \cite{Zhang2019,Zou2020,Vasconcelos2021}. Our results will show that combining the quantile modification with blurring-based modifications is better than both alone and is essential towards obtaining robustness. It is also different to other modifications to non-linearities like in \cite{Xie2020a}, whose Smooth ReLu is used to improve \ac{AT} and is used only during the backward pass. Our QReLu is devised to reduce aliasing and improve robustness as a pure structural component only, separate from \ac{AT}, and is used during both the forward and backward pass.

\subsection{Eliminating aliasing in sub-sampling layers}
\label{sec:eliminating-aliasing-in-subsampling-layers}

Not a true layer as it is commonly understood, but can be thought of as a component of any layer $g$ with a stride parameter $S>1$ by factorizing $g$ into a dense evaluation (stride $1$) followed by a sub-sampling
\begin{equation}
	g_{\text{stride}=S} = g_{\text{stride}=1} *_S \delta
\end{equation}
where $*_S$ denotes convolution with stride $S$, and $\delta$ is the Kronecker delta (identity element for the convolution operation in the discrete domain). The anti-alias treatment necessary is well known from the theory of signal processing and consists of blurring to a normalized frequency $\frac{1}{2S}$ prior to sub-sampling, and is independent of the prior dense operation $g$.

\subsection{Eliminating aliasing in max-pooling layers}
\label{sec:eliminating-aliasing-in-maxpooling-layers}

Pooling layers may introduce aliasing in one of two ways. First, the pooling operation, if it is non-linear. Second, any posterior sub-sampling due to a non-unitary stride. The solution to the second source of aliasing is covered in \cref{sec:eliminating-aliasing-in-subsampling-layers}; its specific application to a max-pool was introduced in \cite{Zhang2019} by the name of \ac{MBP}. The first source of aliasing is tricky to deal with using the tools we have developed, as max-pooling does not have a good small-degree polynomial approximation. Hence, we turn to a heuristic argument based on non-uniform sampling principles.

We re-interpret a max-pooling operation with $\text{kernel size}=\text{stride}=K$ as a non-uniform sampling that chooses one representative for every neighborhood of size $K \times K$. In general, we have an average "sampling rate" of $\frac{1}{K}$, so we have to blur to a normalized frequency $\frac{1}{2K}$ in order to avoid aliasing \cite{ClaudeShannon1949, Maymon2011}. Given the appropriate blurring strength, this anti-aliasing modification is equivalent to the post-filtering of \cite{Vasconcelos2021}, though they arrived to it via a different argument. Whereas their focus was on improving training dynamics with a generalization maximizing objective, ours arises from non-uniform sampling in an adversarial setting. The advantage of our argument is that the appropriate blurring strength is an intrinsic part of it.

Without up-sampling, an anti-aliased pool may be computed as such. The first blur heuristically anti-aliases the pooling operation as just explained, and the second blur anti-aliases the implicit stride (\cref{sec:eliminating-aliasing-in-subsampling-layers}).
\begin{equation}
	\text{AAPool}_{\text{stride}=K} = \text{Blur}_{\frac{1}{2K},\text{stride}=K} \circ \text{Pool}_{\text{stride}=1} \circ \text{Blur}_{\frac{1}{2K}}
\end{equation}
By allowing up-sampling, we can relax the strength of the first blur just like with the \relu (\cref{sec:reducing-aliasing-in-the-relu}).

% Summary graphic
\begin{figure}
    \centering
    \includegraphics[width=1\linewidth]{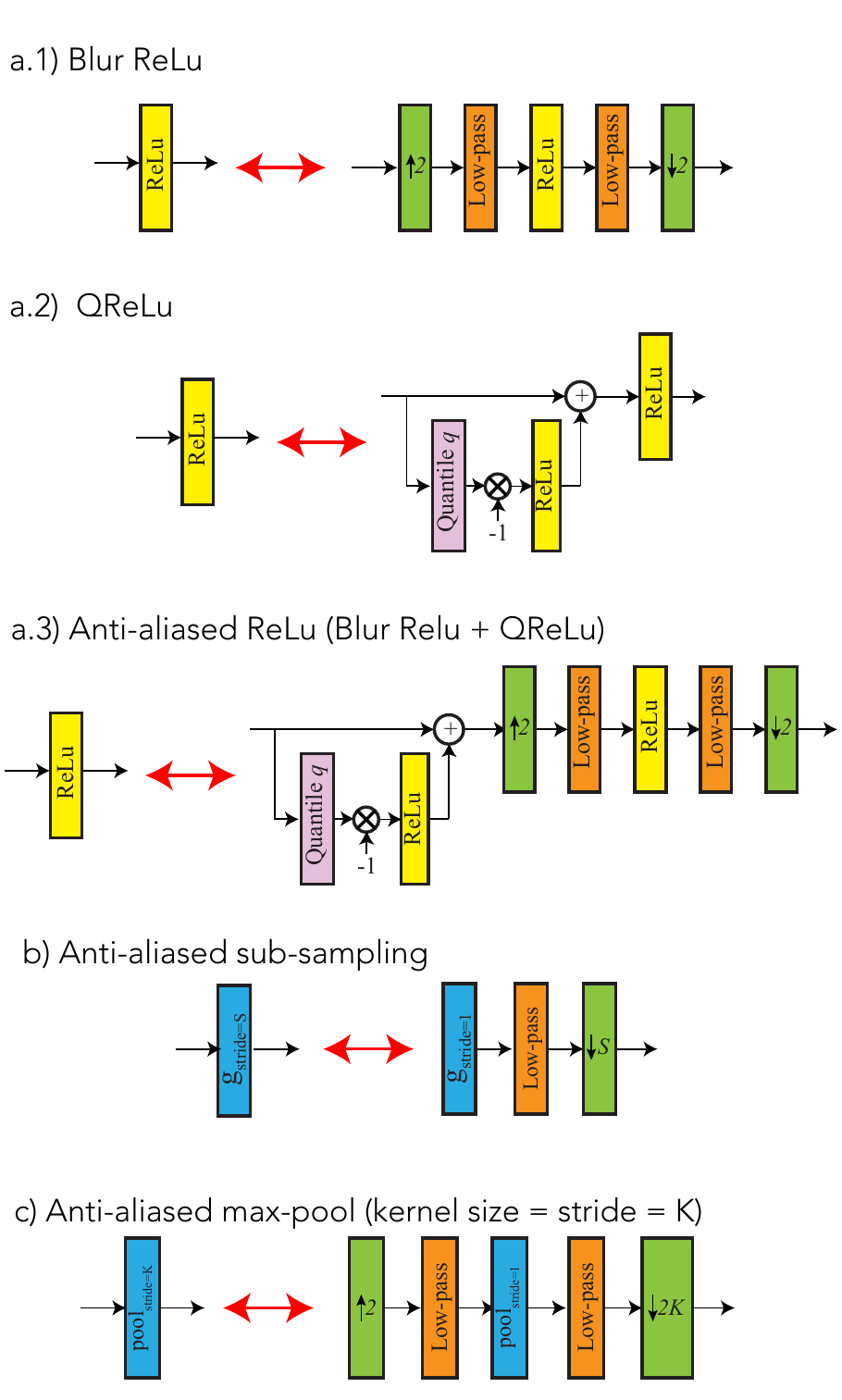}
	\caption{List of anti-aliasing replacements: (a.1) Blur \relu, (a.2) QReLu, (a.3) Anti-aliased ReLu (combining Blur \relu and QReLu), (b) sub-sampling, and (c) max-pooling.}
	\label{figure:layercards}
\end{figure}

\subsection{Practical considerations}

{\bf Practical blur kernels:} In practice, reasonably sized blurring kernels have non-zero width transition bands, so there is an essential trade-off between anti-aliasing strength and quality degradation. In our experiments we have used separable Kaiser filters. As in \cite{Karras2021}, for a given target normalized frequency $\frac{1}{2L}$ we set the following Kaiser parameters: $F_c=\frac{1}{4L}$ to strike a balance between anti-aliasing and image quality degradation, and $F_h \coloneqq \frac{1}{2} - F_c$ to minimize rippling. Furthermore, we set $\texttt{numtaps} \coloneqq \min(\max(2\lfloor s/8 \rfloor + 1, 3), 7)$, where $s$ is the height of the feature map, to reduce boundary artifacts.

{\bf Upsampling vs blurring:} In practice we may use blurring, up-sampling, or a combination of both to perform non-linear transformations without aliasing. The usage of up-sampling reduces the amount of high-frequency information that needs to be destroyed, and improves the accuracy-robustness trade-off. Any up-sampling used for this purpose has a corresponding down-sampling to keep input/output resolutions the same for each layer as seen in \cref{figure:layercards}. For both the \relu and max-pool we used an up-sampling factor of $2$, just like \cite{Karras2021}. Moreover, up-sampling by a higher amount results in quadratic memory costs, so we think this is an acceptable compromise between compute, robustness, and accuracy.

%-------------------------------------------------------------------------
\section{Experiments and results}
\label{sec:experiments-and-results}

Previous work on anti-aliasing such as \cite{Zhang2019,Zou2020,Vasconcelos2021,Karras2021} has focused on improving generalization and model quality. The sole exception is \cite{Grabinski2021}, which acutely observed that aliasing coincided with adversarial vulnerability, and suggested that integration of signal processing concepts into networks was necessary to correct adversarial vulnerability at the root, but left this task to future work. Our contribution is to provide exactly this integration, and propose concrete anti-aliasing modifications as an explicit un-trained structural defense to adversarial attacks. This section analyzes the effect on robustness of the modifications proposed in \cref{sec:methodology}, and compares them to vanilla and \ac{AT} approaches. 

In particular, we observe that simply reducing aliasing already results in more robust classifiers, and combining anti-aliasing with robust training out-performs solo robust training on $L_2$ attacks with no or minimal losses in performance on $L_{\infty}$ attacks.

We start by detailing our experimental settings (\cref{sec:experimental-settings}), most notably the architectures, datasets, and adversarial attacks used for evaluation. Secondly, we investigate the workings of anti-aliasing, specifically with regards to anti-aliasing depth (\cref{sec:accuracy-AAdepth}) and the interplay between the blurring-based modifications and the QReLu (\cref{sec:interaction}). Thirdly, we evaluate anti-aliasing as a defense and compare it to robust training (\cref{sec:accuracy-amplitude}). Lastly, we have a discussion on computational cost (\cref{sec:computecost}).

\subsection{Experimental settings}
\label{sec:experimental-settings}

As is standard in adversarial attacks research, we used the Cifar-10 dataset \cite{Krizhevsky2009} for our experiments. Additionally, we also used the larger TinyImagenet dataset \cite{TinyImagenet} ($64\times64$ images as opposed to Cifar's $32\times32$). With respect to the architectures, we used the simple and light-weight VGG11 and the more common Resnet-50. The \ac{AT} models were trained using PGD with the configuration of \cite{Madry2019}, and a half-clean half-adversarial batch approach.

We evaluate robustness using the white-box gradient attacks FGSM and PGD with varying $L_\infty$ and $L_2$ adversarial strengths (epsilon), 20 steps (for PGD), and their default \cite{Kim2020} configurations otherwise, and consider the attacker successful if the model misclassifies the attacked input. We have elected to use white-box gradient attacks since they are the strongest attacker model, having full information about the network. Furthermore, because our anti-aliasing modifications are fully-differentiable and have the exact same behaviour in training and evaluation we eliminate concerns about gradient masking (otherwise the network would not have trained properly or at all). This means that PGD in particular is a very strongly adapted attack as it has full knowledge of our defense's exact training gradient function, and is thus a very good candidate for benchmarking.

\subsection{Is anti-aliasing at all depths required to achieve robustness?}
\label{sec:accuracy-AAdepth}
\begin{figure*}
    \centering
    \includegraphics[width=1\linewidth]{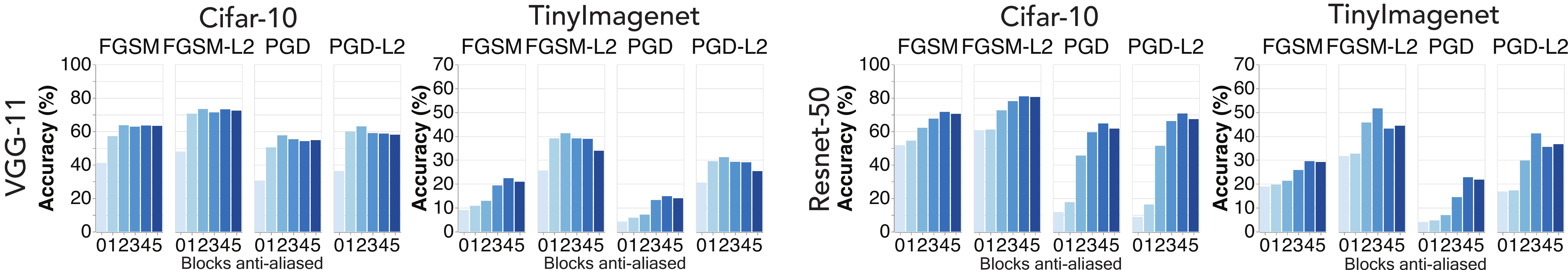}
	\caption{Defense accuracy vs number of blocks anti-aliased on the various attacks at low amplitude (epsilon=$2$). There are diminishing returns in robustness per block anti-aliased, and peak robustness is always obtained at the fourth block or earlier.}
	\label{fig:accuracy_AAdepth}
\end{figure*}

In this section, we investigate whether anti-aliasing all network layers is necessary to obtain robustness, or whether only anti-aliasing a few initial layers, where the feature map is bigger and pixel correlations are image-like, is sufficient. To this end, we split the VGG-11 and the Resnet-50 into five blocks and measure the robustness of a model with only the first $k$ blocks, $0\leq k \leq 5$, anti-aliased, which we denote by AA$(k)$ (Anti-aliasing the first $k$ blocks). $k=0$/AA(0) is equivalent to the vanilla defense \ie, doing nothing.

\cref{fig:accuracy_AAdepth} plots accuracy vs blocks anti-aliased for low-amplitude (epsilon$=2$) attacks. Overall, we see that deeper blocks have a decreasing marginal effect, though the exact magnitudes depend on the attack and to a lower extent the model and dataset. In particular, anti-aliasing just two or three blocks yields maximum or close to maximum robustness in all but one case (Resnet-50+TinyImagenet on PGD).

Additionally, it seems that robustness plateaus faster on Cifar-10 compared to TinyImagenet, on single-step attacks compared to multi-step attacks, and on the VGG-11 compared to the Resnet-50. The first is perhaps due to the smaller sized image; at low resolutions, anti-aliasing just muddles the signal, so it makes sense that this happens earlier on Cifar-10 than TinyImagenet. The second is likely due to the relative strengths of the attack. The third might be because of the residual layer structure of the Resnet-50, which makes image dynamics last longer depth-wise in the network.

\subsection{What is the interaction between the blurring and quantile approaches?}
\label{sec:interaction}
In this section, we investigate the interaction between the two anti-aliasing approaches discussed: the more standard blurring-based modifications, which we bundle together and denote by "Blur" for simplicity, and the novel QReLu. To this end, we test defenses consisting of only one type of approach and compare them to their combination, with the vanilla defense for reference. For this experiment we anti-alias all blocks \ie, Blur+QReLu is equivalent to AA(5).

\begin{figure}
    \centering
    \includegraphics[width=1\linewidth]{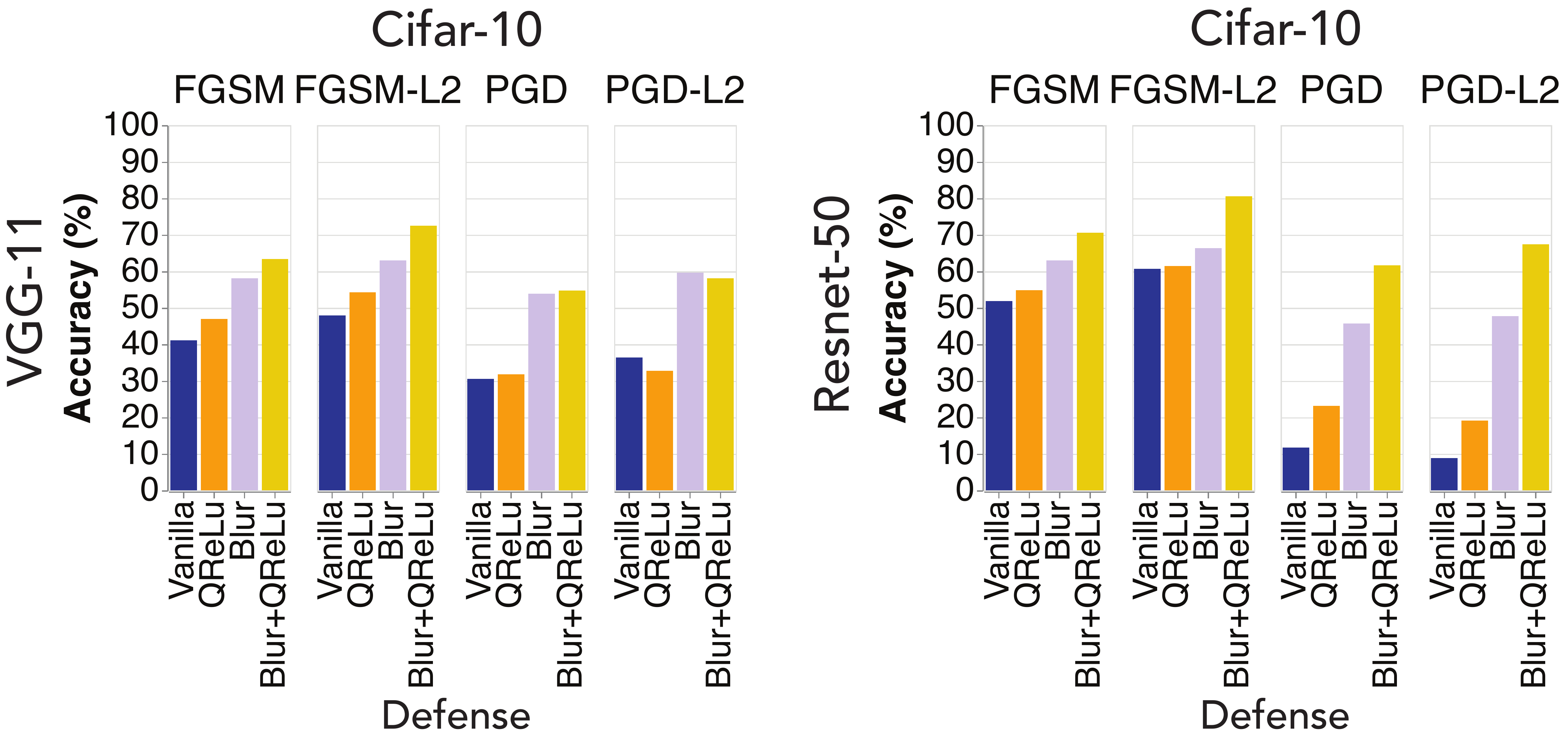}
	\caption{Defense accuracy vs type of approach on the various attacks at low amplitude (epsilon=$2$). Blur+QReLu is the combination of the blurring-based modifications and the QReLu. The two are additive or super-additive, and QReLu is especially effective on the Resnet-50.}
	\label{fig:interaction}
\end{figure}

\cref{fig:interaction} plots accuracy vs defense for low-amplitude (epsilon$=2$) attacks. We observe that Blur alone achieves similar robustness to Blur+QReLu with the VGG-11, but lags signficantly behind it on the Resnet-50. Analogously, the QReLu obtains small robustness gains alone, but is additive or super-additive when combined with Blur, especially on the Resnet-50. Blur+QReLu outperforms Blur in all but one case where it is only slightly below.

\subsection{How effective is anti-aliasing as a defense?}
\label{sec:accuracy-amplitude}

In this section, we compare the robustness of five defenses:
\begin{itemize}
    \item Vanilla: No defense.
    \item Initial Blur: Naive initial blur with [1 4 6 4 1]
    \item AA(5): Anti-aliasing all five blocks of the network.
    \item AT: Adversarial Training with PGD
    \item AT+AA(2): Combining adversarial training with anti-aliasing the first two blocks of the network.
\end{itemize}

%Vanilla: no defense; Initial Blur: a naive initial blur with [1 4 6 4 1]; AT: Adversarial Training with PGD; AA(5): anti-aliasing all five blocks; and AT+AA(2): combining adversarial training with anti-aliasing the first two blocks. 

\begin{figure*}
    \centering
    \includegraphics[width=0.95\linewidth]{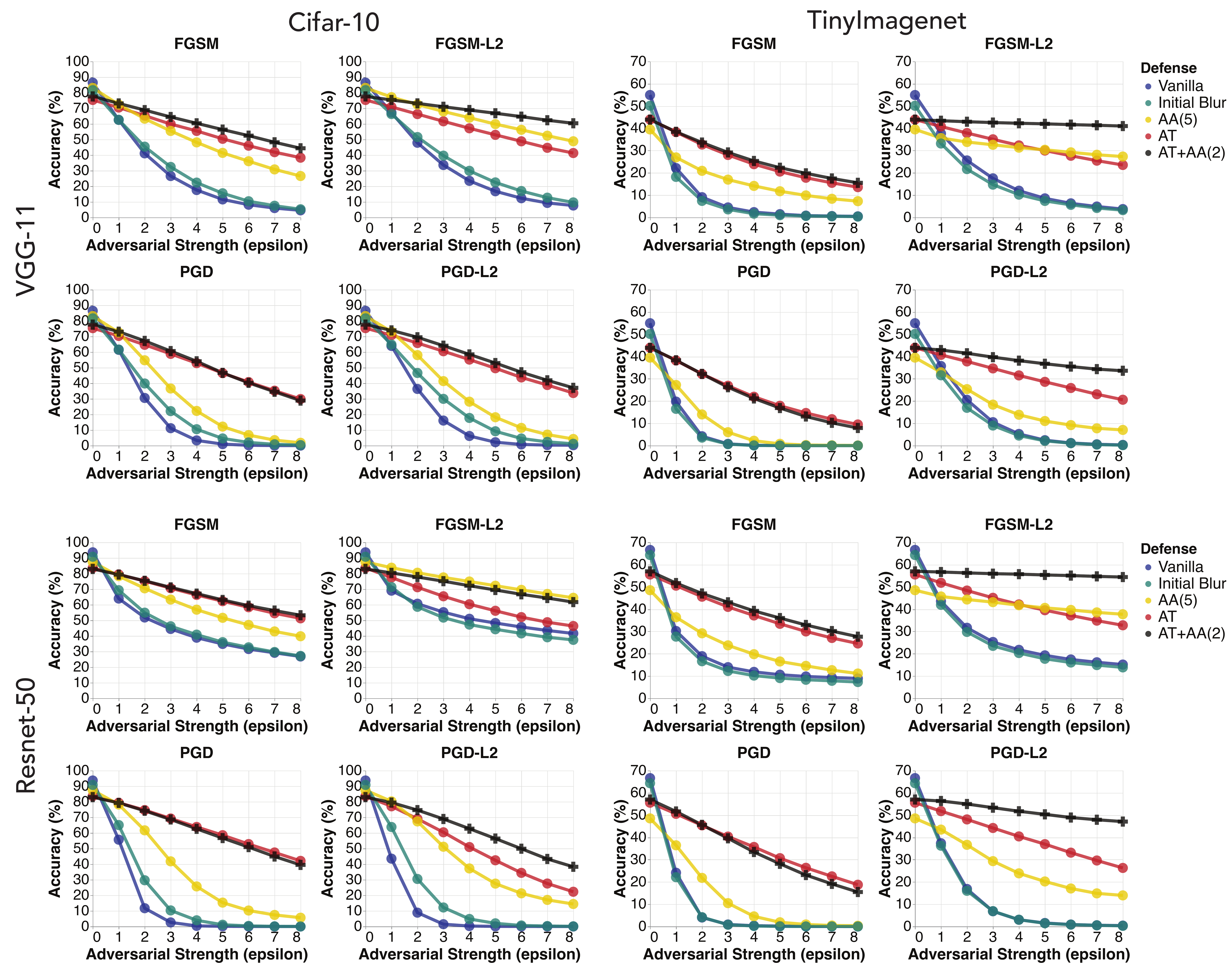}
	\caption{Defense accuracy vs adversarial strength (epsilon) curves for the Vanilla, Initial Blur, AA(5), AT, and AT+AA(2) defenses on the various attacks. Anti-aliasing by itself is already effective against low-amplitude attacks and single-step attacks, especially the $L_2$ variants. Combining anti-aliasing with robust training, AT+AA(2), out-performs solo robust training, AT, on $L_2$ attacks with only minimal losses on $L_{\infty}$ attacks.}
	\label{fig:accuracy_amplitude}
\end{figure*}

\cref{fig:accuracy_amplitude} plots adversarial strength vs accuracy curves for the five defenses against various attacks on each architecture and dataset. The results show that, while not the full picture, aliasing plays a significant role in the vulnerability of vanilla networks to adversarial attacks. 

Our simple anti-aliasing measures derived in \cref{sec:methodology} are sufficient by themselves to increase the robustness of networks significantly for low-amplitude and single-step attacks, especially for the $L_2$ variants. Most notably, AA(5) beats AT on the FGSM $L_2$ attack for all amplitudes on 3 out of the 4 dataset+architecture combinations. Moreover, the relative brittleness of Initial Blur shows that naive blurring approaches are insufficient, which further validates our methodology.

Furthermore, we observe that the AT+AA(2) defense, which combines anti-aliasing with robust training, consistently outperforms the robust training defense AT on $L_2$ attacks, sometimes by a very wide margin such as in Resnet-50+TinyImagenet where both $L_2$ attacks appear completely beaten, while maintaining statistically equal robustness on $L_{\infty}$ attacks. Adding anti-aliasing helps robust training generalize to attacks with a different norm-constraint than the one used for training. Any minimal losses in performance on $L_\infty$ attacks can be explained by recognizing that the robust training is done precisely with PGD, which implies the anti-aliased robust training has a smaller model search space, and additionally some overfitting to the attack might be at play given that AT is robust training with PGD $L_{\infty}$.

%Unfortunately, the effect seems to largely collapse for high-amplitude multi-step attacks. We suspect the higher perturbation and search budget allows the attacker to exploit other effects, which seem to be sufficient to still confound the network, though it is unable to do so without the extra resources.

%Our conclusion is that there exist additional sources of vulnerability besides aliasing, but which require a higher perturbation and search budget to exploit. Unfortunately, unless almost all the different sources of vulnerability are patched, powerful adversaries will be able to confound models. Nevertheless, our results show that aliasing is a large source of vulnerability, and that explainable non-trained structural-only defenses must include an anti-aliasing component.

\subsection{What is the computational cost of anti-aliasing modifications?}
\label{sec:computecost}
\begin{figure}
    \centering
    \includegraphics[width=1\linewidth]{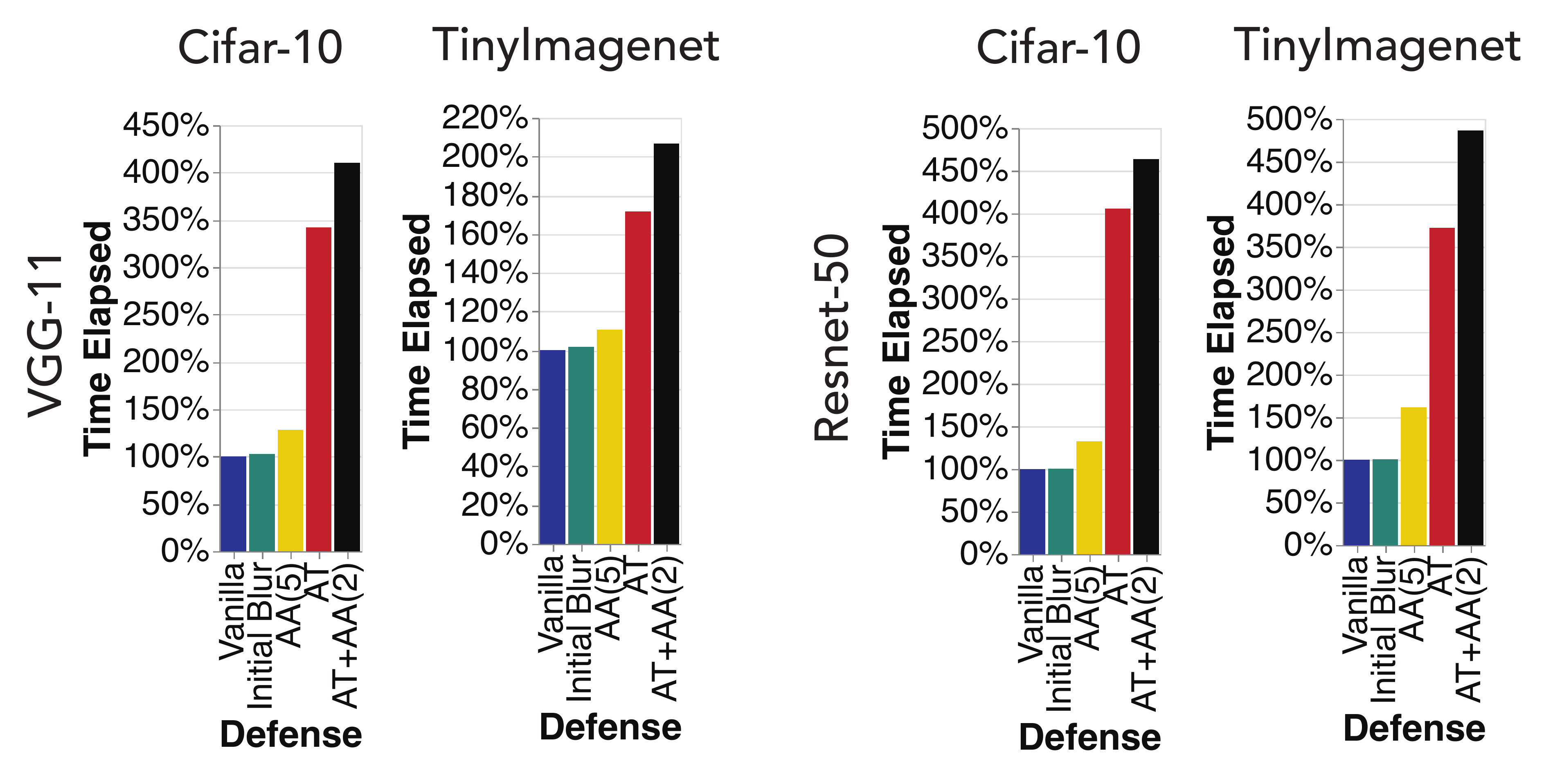}
	\caption{Training time for each defense relative to the Vanilla defense. Anti-aliasing (FA) usually results in a relatively small 10-30\% increase in training time, while robust training is routinely 200-300\% more costly.}
	\label{fig:computecost}
\end{figure}
In this section, we evaluate the computational cost of anti-aliasing defenses compared to robust training. 

\cref{fig:computecost} plots the training time of each defense (with identical training regimes) relative to the Vanilla defense. We observe how anti-aliasing (AA(5)) usually results in a relatively small 10-30\% increase in training time, with the exception of Resnet-50+TinyImagenet which had a 60\% increase. Conversely, robust training routinely has a 200-300\% increase in training time, with the exception of VGG-11+TinyImagenet where it had a 70\% increase. Overall, robust training is routinely 2.5x-3x more expensive than anti-aliasing, which is generally fairly light-weight. The computational cost of combining anti-aliasing with robust training yields smaller metrics along a similar vein, as anti-aliasing two blocks is much cheaper than anti-aliasing five.

\subsection{What are the visual effects of the defenses on the adversarial perturbations?}
\begin{figure}
    \centering
    \includegraphics[width=0.975\linewidth]{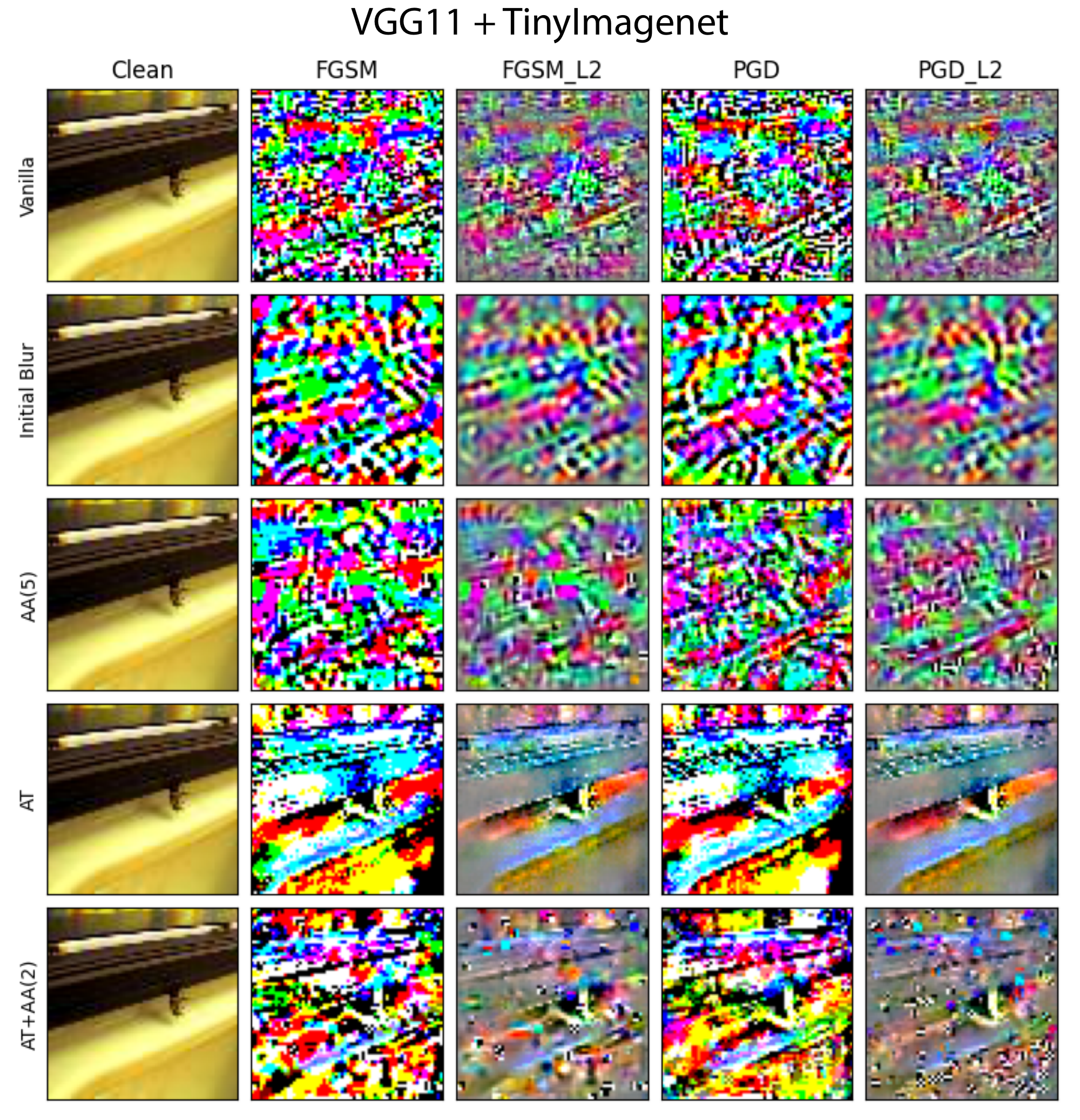} \\
    \vspace{0.6cm}
    \includegraphics[width=0.975\linewidth]{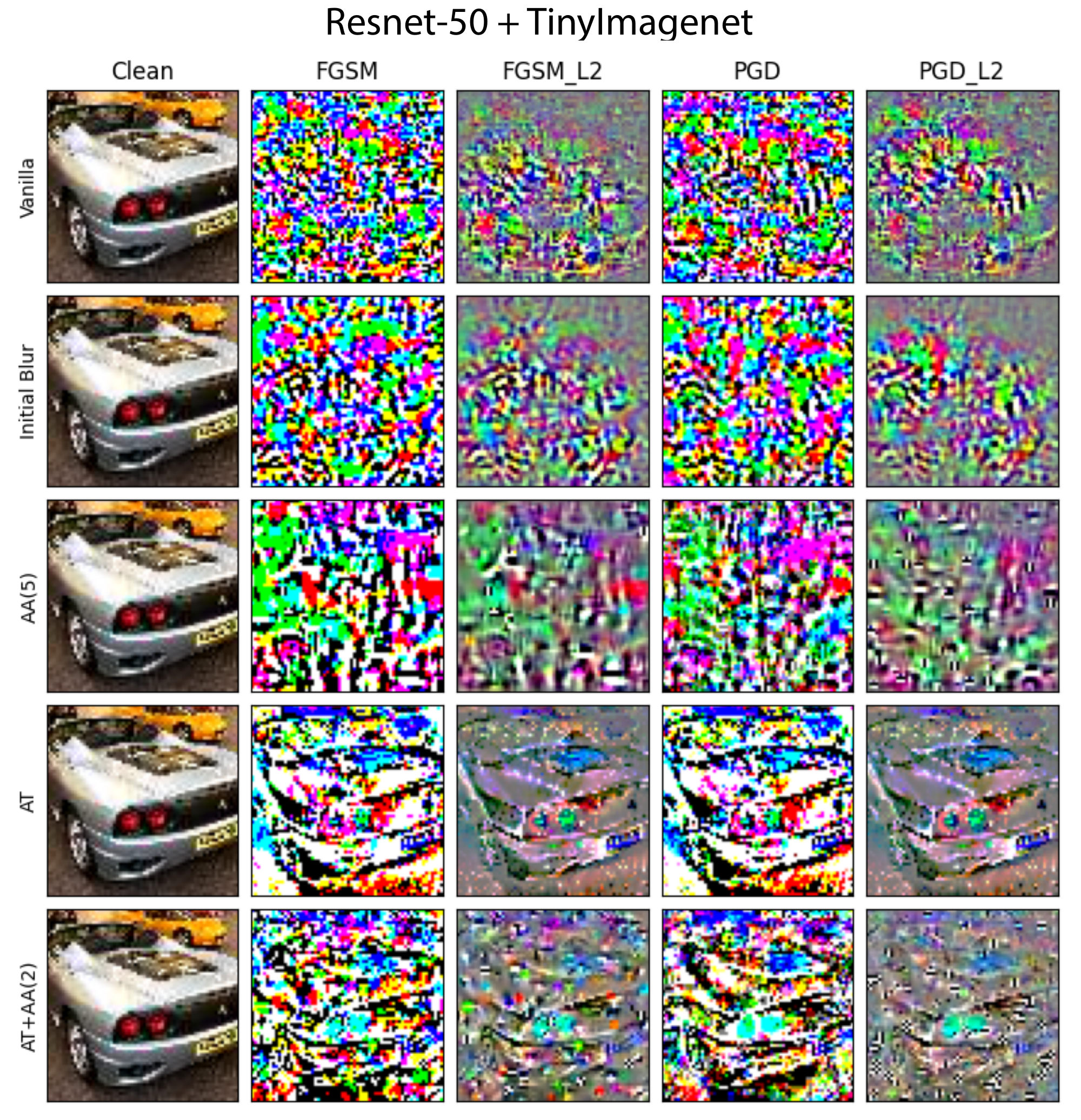}
	\caption{Perturbation examples for the different defenses and attacks at low amplitude (epsilon=$2$) on TinyImagenet. Particularly notable is the resemblance between PGD-AT noises and the clean image. A smaller degree of resemblance (particularly the edges) can be seen between FA perturbations and the clean image.}
	\label{fig:examplegen}
\end{figure}

\cref{fig:examplegen} showcase perturbation examples for the various defenses and attacks at low amplitude (epsilon=$2$). A natural-like attack perturbation shows that the model is more responsive to natural images and less responsive to random white-like noise, which aligns with the intuition for anti-aliasing mentioned in the introduction. Vanilla model perturbations do not resemble the clean image in either edges, coloration or texture. FA model perturbations show noticeable increase in the resemblance of the edges, but not in coloration or texture, compared to Vanilla model perturbations. PGD-AT perturbations show greatly increased resemblance in edges and texture, and slightly in coloration.

%------------------------------------------------------------------------

\section{Conclusions}
\label{sec:conclusions}
Is aliasing in neural networks responsible for their vulnerability to adversarial attacks? The experimental results presented in this paper empirically show that anti-aliasing alone makes networks significantly more robust to any-amplitude single-step attacks and low-amplitude multi-step attacks. Furthermore, combining anti-aliasing with robust training out-performs solo robust training on $L_2$ attacks with no or minimal losses on $L_{\infty}$ attacks.
\paragraph{Broader impact:} Vulnerability to attacks by current classifiers hinders their applicability to many domains. Advances in the understanding of the sources of vulnerability is important to improve classifiers and to open the door to applications where reliability is key.

%%%%%%%%% REFERENCES
{\small
\bibliographystyle{ieee_fullname}
\bibliography{citations}
}

%%%%%%%%% Supplementary Material
\appendix
\onecolumn

\section{Supplementary Material}

The supplementary material consists of an expanded version of \cref{fig:toy_example} that includes Fourier Transforms, shedding more light on how the aliasing-based attack works, the results from \cref{fig:accuracy_amplitude} in tabular form, and a rigorous treatment of the claims made in \cref{sec:reducing-aliasing-in-the-relu}.

%------------------------------------------------------------------------

\subsection{Fourier-based analysis of aliasing-based attack}

\renewcommand{\thefigure}{A1}
\begin{figure*}[h]
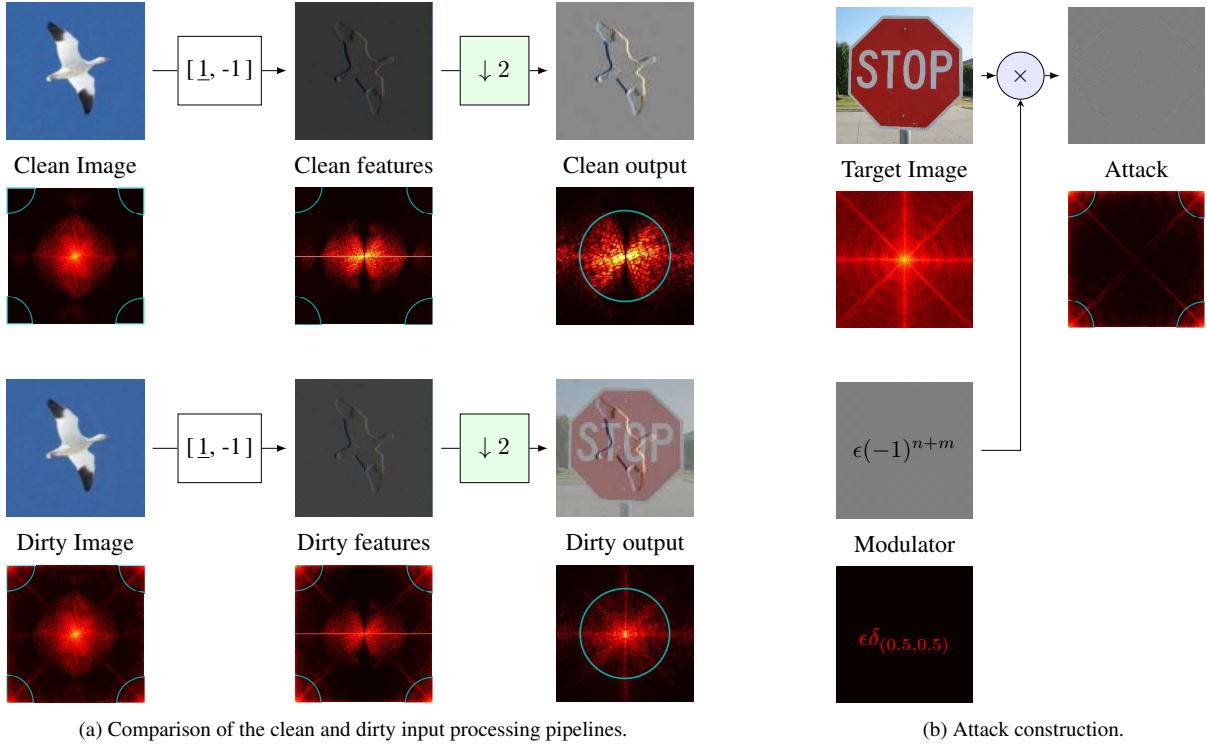

\centering
\begin{subfigure}[b]{0.60\textwidth}
  \centering
  \resizebox{.9\linewidth}{!}{\input{figures/toy_example/pipeline_fourier}}
  \caption{Comparison of the clean and dirty input processing pipelines.}
\end{subfigure}
~~~~~~~~\begin{subfigure}[b]{0.33\textwidth}
  \centering
  \resizebox{.9\linewidth}{!}{\input{figures/toy_example/construction_fourier}}
  \caption{Attack construction.}
\end{subfigure}
\caption{Extension of \cref{fig:toy_example} including 2D Fourier Transforms (FT). We observe how the attacker, shown in (b), infiltrates the high frequency regions of the input image, shown in the corners of the FT plots. The sub-sampling operation causes high frequencies to occupy, or "fold into", the lower frequency space, overlapping with the clean signal and turning the attack from invisible to visible. This radical change in appearance after a sub-sampling operation is what is meant in signal processing by aliasing.}
\label{fig:toy_example_fourier}
\end{figure*}

\cref{fig:toy_example_fourier} is an expansion of \cref{fig:toy_example} that includes Fourier Transforms (constructed to have the center position match the 2D frequency $(0, 0)$). From the Fourier Transforms we observe how in the beginning, the attack lives at the edges of the frequency space, which correspond to high frequencies, and is separate from the signal living in the center, which corresponds to the lower-middle frequencies. However, after sub-sampling, the previously high frequencies start occupying the low frequency space, making the attack visible and corrupting the signal. This latter step is also known as "frequency folding", describing how the high frequencies "fold into" the lower frequency space and which is an interpretation of the aliasing in the frequency domain.

Essentially, the concept of aliasing describes how an insufficient sampling rate can make high frequencies and low frequencies look the same. In this adversarial scenario, the attack, previously "invisible" to the human eye by virtue of being a high frequency signal, becomes visible after the sub-sampling. Removing the attack via anti-aliasing prior to the sub-sampling is the only way to protect the integrity of the signal. Here we employ the domain knowledge whereby natural images are almost completely a lower-middle frequency signal, and anti-aliasing eliminates potentially dangerous content in the high-frequency space without major disruption to the clean image.

%------------------------------------------------------------------------

\subsection{Tabular results}

\subsubsection{Figure 6: Defense accuracy vs adversarial strength (epsilon) curve}

\begin{table}[H]
    \centering
    \begin{minipage}{.48\textwidth}
    \begin{adjustbox}{max width=\textwidth}
        % Please add the following required packages to your document preamble:
% \usepackage{booktabs}
% \usepackage{multirow}
\begin{tabular}{@{}rrrrrr@{}}
\toprule
\textbf{}                     & \textbf{Attack}    & \textbf{FGSM}  & \textbf{FGSM-L2} & \textbf{PGD}   & \textbf{PGD-L2} \\ \midrule
\textbf{Defense}              & \textbf{Epsilon} & \textbf{}      & \textbf{}        & \textbf{}      & \textbf{}       \\ \midrule
\multirow{4}{*}{Vanilla}      & 0                  & \textbf{86.48} & \textbf{86.48}   & \textbf{86.48} & \textbf{86.48}  \\
                              & 2                  & 41.10          & 47.91            & 30.58          & 36.37           \\
                              & 4                  & 17.59          & 23.48            & 3.36           & 6.20            \\
                              & 8                  & 4.57           & 7.66             & 0.03           & 0.28            \\ \midrule
\multirow{4}{*}{Initial Blur} & 0                  & 81.61          & 81.61            & 81.61          & 81.61           \\
                              & 2                  & 45.41          & 51.58            & 39.86          & 46.61           \\
                              & 4                  & 22.35          & 29.91            & 10.54          & 17.77           \\
                              & 8                  & 5.26           & 9.65             & 0.35           & 1.19            \\ \midrule
\multirow{4}{*}{AA(5)}        & 0                  & 83.13          & 83.13            & 83.13          & 83.13           \\
                              & 2                  & 63.35          & 72.47            & 54.75          & 58.06           \\
                              & 4                  & 48.15          & 63.97            & 22.27          & 28.20           \\
                              & 8                  & 26.58          & 48.86            & 1.84           & 4.28            \\ \midrule
\multirow{4}{*}{AT}           & 0                  & 75.35          & 75.35            & 75.35          & 75.35           \\
                              & 2                  & 65.30          & 66.29            & 64.67          & 65.80           \\
                              & 4                  & 55.49          & 57.18            & 52.90          & 55.19           \\
                              & 8                  & 38.34          & 41.32            & \textbf{29.90} & 33.74           \\ \midrule
\multirow{4}{*}{AT+AA(2)}     & 0                  & 77.62          & 77.62            & 77.62          & 77.62           \\
                              & 2                  & \textbf{68.85} & \textbf{73.17}   & \textbf{67.17} & \textbf{69.50}  \\
                              & 4                  & \textbf{60.39} & \textbf{68.89}   & \textbf{54.11} & \textbf{58.51}  \\
                              & 8                  & \textbf{44.38} & \textbf{60.41}   & 28.90          & \textbf{37.07}  \\ \bottomrule
\end{tabular}
    \end{adjustbox}
    \caption{\cref{fig:accuracy_amplitude} tabular results for VGG-11 + Cifar-10}
     \label{table:accuracy_amplitude:vgg11-cifar10}
    \end{minipage}%
    \hspace{0.3cm}
    \begin{minipage}{.48\textwidth}
    \begin{adjustbox}{max width=\textwidth}
        % Please add the following required packages to your document preamble:
% \usepackage{booktabs}
% \usepackage{multirow}
\begin{tabular}{@{}rrrrrr@{}}
\toprule
\textbf{}                     & \textbf{Attack}    & \textbf{FGSM}  & \textbf{FGSM-L2} & \textbf{PGD}   & \textbf{PGD-L2} \\ \midrule
\textbf{Defense}              & \textbf{Epsilon} & \textbf{}      & \textbf{}        & \textbf{}      & \textbf{}       \\ \midrule
\multirow{4}{*}{Vanilla}      & 0                  & \textbf{54.97} & \textbf{54.97}   & \textbf{54.97} & \textbf{54.97}  \\
                              & 2                  & 9.11           & 25.61            & 4.23           & 20.58           \\
                              & 4                  & 2.40           & 12.02            & 0.15           & 5.25            \\
                              & 8                  & 0.50           & 3.77             & 0.02           & 0.33            \\ \midrule
\multirow{4}{*}{Initial Blur} & 0                  & 50.15          & 50.15            & 50.15          & 50.15           \\
                              & 2                  & 7.44           & 21.72            & 3.60           & 16.98           \\
                              & 4                  & 1.70           & 10.25            & 0.09           & 4.52            \\
                              & 8                  & 0.39           & 3.29             & 0.01           & 0.21            \\ \midrule
\multirow{4}{*}{AA(5)}        & 0                  & 39.52          & 39.52            & 39.52          & 39.52           \\
                              & 2                  & 20.94          & 33.87            & 14.01          & 25.37           \\
                              & 4                  & 14.17          & 31.34            & 2.20           & 13.87           \\
                              & 8                  & 7.31           & 27.35            & 0.09           & 7.06            \\ \midrule
\multirow{4}{*}{AT}           & 0                  & 43.95          & 43.95            & 43.95          & 43.95           \\
                              & 2                  & 32.73          & 37.90            & 32.15          & 37.80           \\
                              & 4                  & 23.92          & 32.35            & \textbf{21.96} & 31.53           \\
                              & 8                  & 13.61          & 23.56            & \textbf{9.51}  & 20.61           \\ \midrule
\multirow{4}{*}{AT+AA(2)}     & 0                  & 43.92          & 43.92            & 43.92          & 43.92           \\
                              & 2                  & \textbf{33.64} & \textbf{42.98}   & \textbf{32.21} & \textbf{41.47}  \\
                              & 4                  & \textbf{25.37} & \textbf{42.29}   & 21.19          & \textbf{38.16}  \\
                              & 8                  & \textbf{15.55} & \textbf{41.03}   & 7.92           & \textbf{33.62}  \\ \bottomrule
\end{tabular}
    \end{adjustbox}
    \caption{Tabular results for \cref{fig:accuracy_amplitude} for VGG-11 + TinyImagenet}
    \label{table:accuracy_amplitude:vgg11-tinyimagenet}
    \end{minipage}
\end{table}

\begin{table}[H]
    \centering
    \begin{minipage}{.48\textwidth}
    \begin{adjustbox}{max width=\textwidth}
        % Please add the following required packages to your document preamble:
% \usepackage{booktabs}
% \usepackage{multirow}
\begin{tabular}{@{}rrrrrr@{}}
\toprule
\textbf{}                     & \textbf{Attack}    & \textbf{FGSM}  & \textbf{FGSM-L2} & \textbf{PGD}   & \textbf{PGD-L2} \\ \midrule
\textbf{Defense}              & \textbf{Epsilon} & \textbf{}      & \textbf{}        & \textbf{}      & \textbf{}       \\ \midrule
\multirow{4}{*}{Vanilla}      & 0                  & \textbf{93.63} & \textbf{93.63}   & \textbf{93.63} & \textbf{93.63}  \\
                              & 2                  & 51.83          & 60.67            & 11.73          & 8.88            \\
                              & 4                  & 38.87          & 51.01            & 0.39           & 0.29            \\
                              & 8                  & 26.83          & 41.59            & 0.00           & 0.00            \\ \midrule
\multirow{4}{*}{Initial Blur} & 0                  & 90.80          & 90.80            & 90.80          & 90.80           \\
                              & 2                  & 54.93          & 58.68            & 29.66          & 30.49           \\
                              & 4                  & 40.84          & 47.29            & 3.97           & 4.83            \\
                              & 8                  & 27.19          & 37.49            & 0.03           & 0.11            \\ \midrule
\multirow{4}{*}{AA(5)}        & 0                  & 87.44          & 87.44            & 87.44          & 87.44           \\
                              & 2                  & 70.53          & \textbf{80.54}   & 61.67          & 67.38           \\
                              & 4                  & 56.85          & \textbf{74.91}   & 25.71          & 37.23           \\
                              & 8                  & 39.81          & \textbf{64.46}   & 5.65           & 14.43           \\ \midrule
\multirow{4}{*}{AT}           & 0                  & 83.58          & 83.58            & 83.58          & 83.58           \\
                              & 2                  & 75.18          & 71.23            & \textbf{74.66} & 69.07           \\
                              & 4                  & 66.30          & 60.34            & \textbf{63.85} & 50.96           \\
                              & 8                  & 51.29          & 46.36            & \textbf{42.08} & 22.28           \\ \midrule
\multirow{4}{*}{AT+AA(2)}     & 0                  & 82.95          & 82.95            & 82.95          & 82.95           \\
                              & 2                  & \textbf{75.36} & 77.78            & 74.20          & \textbf{74.64}  \\
                              & 4                  & \textbf{67.15} & 72.20            & 62.63          & \textbf{62.73}  \\
                              & 8                  & \textbf{53.26} & 61.70            & 39.50          & \textbf{38.29}  \\ \bottomrule
\end{tabular}
    \end{adjustbox}
    \caption{Tabular results for \cref{fig:accuracy_amplitude} for Resnet-50 + Cifar-10}
     \label{table:accuracy_amplitude:resnet50-cifar10}
    \end{minipage}%
    \hspace{0.3cm}
    \begin{minipage}{.48\textwidth}
    \begin{adjustbox}{max width=\textwidth}
        % Please add the following required packages to your document preamble:
% \usepackage{booktabs}
% \usepackage{multirow}
\begin{tabular}{@{}rrrrrr@{}}
\toprule
\textbf{}                     & \textbf{Attack}    & \textbf{FGSM}  & \textbf{FGSM-L2} & \textbf{PGD}   & \textbf{PGD-L2} \\ \midrule
\textbf{Defense}              & \textbf{Epsilon} & \textbf{}      & \textbf{}        & \textbf{}      & \textbf{}       \\ \midrule
\multirow{4}{*}{Vanilla}      & 0                  & \textbf{66.61} & \textbf{66.61}   & \textbf{66.61} & \textbf{66.61}  \\
                              & 2                  & 18.93          & 31.67            & 4.04           & 16.85           \\
                              & 4                  & 11.87          & 21.76            & 0.36           & 3.03            \\
                              & 8                  & 8.90           & 15.15            & 0.01           & 0.40            \\ \midrule
\multirow{4}{*}{Initial Blur} & 0                  & 64.33          & 64.33            & 64.33          & 64.33           \\
                              & 2                  & 16.65          & 29.80            & 4.19           & 16.02           \\
                              & 4                  & 10.21          & 20.33            & 0.23           & 3.05            \\
                              & 8                  & 7.31           & 13.88            & 0.00           & 0.30            \\ \midrule
\multirow{4}{*}{AA(5)}        & 0                  & 48.53          & 48.53            & 48.53          & 48.53           \\
                              & 2                  & 29.19          & 44.36            & 21.83          & 36.67           \\
                              & 4                  & 19.81          & 41.78            & 4.55           & 23.86           \\
                              & 8                  & 11.13          & 37.74            & 0.34           & 13.96           \\ \midrule
\multirow{4}{*}{AT}           & 0                  & 55.57          & 55.57            & 55.57          & 55.57           \\
                              & 2                  & 45.62          & 48.38            & 45.41          & 48.07           \\
                              & 4                  & 37.19          & 42.15            & \textbf{35.71} & 40.44           \\
                              & 8                  & 24.61          & 32.78            & \textbf{18.75} & 26.29           \\ \midrule
\multirow{4}{*}{AT+AA(2)}     & 0                  & 57.01          & 57.01            & 57.01          & 57.01           \\
                              & 2                  & \textbf{47.07} & \textbf{56.30}   & \textbf{45.62} & \textbf{54.98}  \\
                              & 4                  & \textbf{39.24} & \textbf{55.73}   & 33.44          & \textbf{51.74}  \\
                              & 8                  & \textbf{27.66} & \textbf{54.43}   & 15.48          & \textbf{47.10}  \\ \bottomrule
\end{tabular}
    \end{adjustbox}
    \caption{Tabular results for \cref{fig:accuracy_amplitude} for Resnet-50 + TinyImagenet}
    \label{table:accuracy_amplitude:resnet50-tinyimagenet}
    \end{minipage}
\end{table}

%------------------------------------------------------------------------

\subsection{Bound on aliasing caused by arbitrary activations}

\begin{definition}[Definition of a signal]
\label{definition:signal}
We define a \textit{signal} to be a continuous, piece-wise smooth function $z: \reals^2 \rightarrow \reals$ that is 1-periodic in all its arguments.
\end{definition}

\begin{definition}[Band-limited signals]
\label{definition:bandlimited_signals}
We define the band-limit of a signal as the smallest positive integer $B$ such that the Fourier Series of the signal has coefficient $0$ for frequencies above $B$. Equivalently, signals with band-limit $B$ are those who have no (multi-dimensional) frequency content above $B$.
\end{definition}

\begin{proposition}[Aliasing bound for point-wise functions]\ \\
\label{propostion:aliasing_bound_nonpolynomial}
Let $z$ be a signal with finite bandlimit $B$. Let $\mu(z)$ be the probability measure induced by $z$ and let $\{p_k\}_{k\geq 0}$ be the orthonormal polynomial family induced by $\mu(z)$. Let $\varphi: \reals \rightarrow \reals$ be an arbitrary point-wise function that is square-integrable with respect to $\mu(z)$. Then for any sampling rate $s > 2B$, let $Z$ be the sampling of $z$ with sampling rate $s$, let $r \coloneqq \max \{k \in \integers : s > 2kB\}$, and let $\varphi_r$ be the projection of $\varphi$ onto $\text{Span}\left(\{p_k\}_{0\leq k\leq r}\right)$.

\noindent We have that the aliasing error of the discrete computation $\varphi(Z)$ is bounded by twice the approximation error of $\varphi_r$
\begin{alignat}{2}
    \norm{\varphi(z)-\phi_s(\varphi(Z)))}^2 &= \intunitsq\intunitsq\left|\varphi(z)-\phi_s(\varphi(Z))\right|^2 \\
    &\leq 2\norm{\varphi - \varphi_r}^2_{\mu(z)} \coloneqq \int_{-\infty}^\infty|\varphi - \varphi_{r}|^2d\mu(z)
\end{alignat}
where $\phi_s$ is the ideal sinc interpolator from signal processing.
\begin{proof}
Let $\{c_{k,l}\}_{k,l\in\integers}$ and $\{a_{k,l}\}_{k,l\in\integers}$ be the Fourier Series of $\varphi(z)$ and $\varphi_r(z)$ respectively. The Shannon-Nyquist Sampling Theorem \cite{ClaudeShannon1949} gives the following bound for the aliasing error in terms of the Fourier Series
\begin{alignat}{2}
    \norm{\varphi(z)-\phi_s(\varphi(Z)}^2 &\coloneqq \intunitsq\intunitsq\left|\varphi(z)-\phi_s(\varphi(Z))\right|^2 \\ &\leq 2\sum_{k,l \notin \openzinter{-\frac{s}{2}}{\frac{s}{2}}{2}}|c_{k,l}|^2
\label{equation:aliasing_bound_nonpolynomial:ineq1}
\end{alignat}
Since $\varphi_r(z)$ has band-limit $rB$ by virtue of $\varphi_r$ being a polynomial of degree $r$, and furthermore $s > 2rB$, we have that $a_{k,l} = 0$ for $k,l \notin \openzinter{-\frac{s}{2}}{\frac{s}{2}}{2}$ and thus
\begin{alignat}{2}
    \sum_{k,l \notin \openzinter{-\frac{s}{2}}{\frac{s}{2}}{2}}|c_{k,l}|^2 = \sum_{k,l \notin \openzinter{-\frac{s}{2}}{\frac{s}{2}}{2}}|c_{k,l}-a_{k,l}|^2
    \leq \sum_{k,l\in\integers}|c_{k,l}-a_{k,l}|^2
\label{equation:aliasing_bound_nonpolynomial:ineq2}
\end{alignat}
then, using Parseval's Theorem and that $\varphi,\varphi_r$ are pointwise functions
\begin{alignat}{2}
\sum_{k,l\in\integers}|c_{k,l}-a_{k,l}|^2 &= \norm{\varphi(z) - \varphi_r(z)}^2 \\
&= \intunitsq\intunitsq |\varphi(z)(x,y) - \varphi_r(z)(x,y)|^2dxdy \\
&= \intunitsq\intunitsq |\varphi(z(x,y)) - \varphi_r(z(x,y))|^2dxdy
\label{equation:aliasing_bound_nonpolynomial:ineq3}
\end{alignat}
which by definition of $\mu(z)$ is equal to
\begin{alignat}{2}
\intunitsq\intunitsq |\varphi(z(x,y)) - \varphi_r(z(x,y))|^2dxdy
&= \int_{-\infty}^\infty |\varphi - \varphi_r|^2d\mu(z) \\
&= \norm{\varphi - \varphi_r}^2_{\mu(z)}
\label{equation:aliasing_bound_nonpolynomial:ineq4}
\end{alignat}
Finally, combining \cref{equation:aliasing_bound_nonpolynomial:ineq1,equation:aliasing_bound_nonpolynomial:ineq2,equation:aliasing_bound_nonpolynomial:ineq3,equation:aliasing_bound_nonpolynomial:ineq4} we obtain the desired result
\begin{alignat}{2}
\norm{\varphi(z)-\phi_s(\varphi(Z))}^2 &\coloneqq \intunitsq\intunitsq\left|\varphi(z)-\phi_s(\varphi(Z))\right|^2 \\ &\leq 2\sum_{k,l \notin \openzinter{-\frac{s}{2}}{\frac{s}{2}}{2}}|c_{k,l}|^2 \\
&\leq 2\norm{\varphi - \varphi_r}^2_{\mu(z)}
\end{alignat}
which completes the proof.
\end{proof}
\end{proposition}

\begin{corollary}
\label{corollary:bounded_alias_nonpoly_activations}
Since $\varphi$ is square-integrable with respect to $\mu(z)$, we have that
\begin{equation}
    \lim_{s\rightarrow \infty} \norm{\varphi - \varphi_{r(s)}}^2_{\mu_z} = 0
\end{equation}
where $r(s) \coloneqq \max \{k \in \integers : s > 2kB\}$. Consequently, we have proven that as the sampling rate $s$ grows the aliasing error approaches zero.
\end{corollary}

\begin{observation}
Since in modern architectures Batch Normalization layers always immediately precede activation layers, we know that in most practical examples we will have $\mu(z) = \mathcal{N}(0, 1)$, the standard Normal Distribution with mean $0$ and variance $1$. The orthogonal polynomial family induced by $\mathcal{N}(0, 1)$ is the well-known probabilist's Hermite polynomials. Knowing $\mu(z)$ allows us to obtain a number for the suitable sampling rate.
\end{observation}

\begin{corollary}[Reducing aliasing in the ReLu]
We apply \cref{propostion:aliasing_bound_nonpolynomial} to the case of the ReLu.

The ReLu function $\max(z,0)$ is very well approximated (cosine similarity of $0.9907$) on $\mathcal{N}(0,1)$ by the 4-th degree polynomial
\begin{equation}
    p(z) = \frac{z}{2}-\frac{z^{4}-18z^{2}-9}{24\sqrt{2\pi}}
\end{equation}
Hence, \cref{propostion:aliasing_bound_nonpolynomial} implies a necessary sampling rate $s$ with $s > 4(2B)$, where $B$ is the band-limit of the feature map pre-ReLu, to compute the ReLu with little aliasing.
\end{corollary}

\end{document}